\documentclass[runningheads]{llncs}
\usepackage{graphicx}
\usepackage{graphicx}

\usepackage{booktabs}       
\usepackage{amsfonts}       
\usepackage{nicefrac}       
\usepackage{microtype}      
\usepackage{xcolor}         
\usepackage{mathrsfs}  
\usepackage{bm}
\usepackage{amssymb,amsmath}

\usepackage{subcaption}
\usepackage{wrapfig}
\usepackage{picinpar}
\usepackage{cutwin}
\usepackage{graphicx,multicol}
\usepackage{algorithm}  
\usepackage{algorithmic}
\usepackage{stmaryrd}

\usepackage{multirow}
\usepackage{algorithm}  
\usepackage{algorithmic}
\usepackage{amsfonts}
\usepackage{amsfonts,amssymb} 
\usepackage{verbatim}
\usepackage{url}
\usepackage{enumitem}
\usepackage{hyperref}
\begin{document}
\title{LaksNet: an end-to-end deep learning model for self-driving cars in Udacity simulator}

%
\titlerunning{LaksNet: an end-to-end deep learning model for self-driving cars}

%
%


\author{Lakshmikar R. Polamreddy\inst{1} \and Youshan Zhang\inst{2}}
\authorrunning{L Polamreddy et al.}
\institute{Yeshiva University, New York City, NY 10016, USA,\\
\email{lpolamre@mail.yu.edu}
\and
Yeshiva University, New York City, NY 10016, USA,\\
\email{youshan.zhang@yu.edu}
}
\maketitle              
\begin{abstract}

  The majority of road accidents occur because of human errors, including distraction, recklessness, and drunken driving. One of the effective ways to overcome this dangerous situation is by implementing self-driving technologies in vehicles. In this paper, we focus on building an efficient deep-learning model for self-driving cars. We propose a new and effective convolutional neural network model called `LaksNet' consisting of four convolutional layers and two fully connected layers. We conduct extensive experiments using our LaksNet model with the training data generated from the Udacity simulator. Our model outperforms many existing pre-trained ImageNet and NVIDIA models in terms of the duration of the car for which it drives without going off the track on the simulator.

\end{abstract}
\section{Introduction}

WHO (World Health Organization)~\cite{WHO} reported that approximately 1.3 million people die each year as a result of road traffic crashes. To reduce the accident rate, We could exploit various machine learning algorithms to drive a vehicle without human intervention. End-to-end machine learning models can be built and trained on huge sets of available data in the form of images generated by sensors and cameras. Then, these trained models will drive the vehicles in such a way as to minimize accidents. As we have not yet collected the real-time data, we have made use of the simulation environment developed by Udacity for its nano degree program on self-driving cars. We utilized the Udacity platform to generate training data and to test the performance of our model.


Previous work proposed to develop deep learning models for end-to-end control of a self-driving car, and they mostly utilized the NVIDIA model architecture~\cite{bojarski2016end} and also never compared its performance with the available pre-trained ImageNet models. Some scholars proposed a simpler architecture than that of NVIDIA but could not establish that their model performance was better by comparing the results. In addition, they did not measure the performance of the models in terms of the time the car ran on the track. Our proposed work overcomes the aforementioned limitations.

The major focus of our research is to first check the performance of the NVIDIA model and develop a better model in terms of predicting the steering angles of a self-driving car. We utilize the Udacity simulator for driving the car in autonomous mode on the track. We first train the NVIDIA model with our dataset, then measure its performance. As mentioned in the results section~\ref{sec:results}, it can drive the car for 120 seconds without going off the track. Secondly, we investigate if any of the existing pre-trained ImageNet models would show a better performance than that of the NVIDIA model. Finally, we develop a novel model that achieves the following two objectives. Firstly, our model achieves state-of-the-art performance. Secondly, we build a model with a smaller number of convolutional and fully connected layers so that it requires a smaller number of parameters for training. 

The remainder of the paper is organized as follows. Section~\ref{sec:rel} covers the related work of autonomous driving. Section~\ref{sec:methods} explains the proposed LaksNet model architecture and how it differs from the NVIDIA model. Section~\ref{sec:results and datasets} presents the training and testing datasets of all models and their results. Section~\ref{sec:disc} discusses how hyper-parameter tuning was carried out to arrive at our model architecture. Section~\ref{sec:concl} captures concluding remarks of the proposed work and scope for future work.

\section{Related Work}\label{sec:rel}

Khan et al.~\cite{khan2021autonomous} presented the pros and cons of the implementation of autonomous vehicles. They discussed the benefits such as safety, congestion, traffic management, and the adoption of autonomous vehicle technology by various sectors like mining, freight transportation, and the military industry. Shalev et al.~\cite{shalev2017formal} proposed a white-box, interpretable, mathematical model for safety assurance called Responsibility-Sensitive Safety (RSS) for self-driving cars and designed a system that adhered to safety assurance requirements and was scalable to millions of cars. Ko et al.~\cite{ko2021key} proposed a method for key points estimation and point instance segmentation for lane detection called Point Instance Network (PINet), which can localize the drivable area on the road. Pan et al.~\cite{pan2017virtual} proposed a novel realistic translation network that could be trained in a virtual environment similar to the real world. They concluded that by using synthetic real images as training data in reinforcement learning, the agent generalizes better in a real environment than pure training environment with virtual data or domain randomization. Wu et al.~\cite{wu2022yolop} proposed a YOLOP (You Only Look Once for Panoptic Driving Perception) model, a high-precision and real-time perception system that can assist the vehicle in making reasonable decisions while driving. The network could perform traffic object detection, drivable area segmentation, and lane detection simultaneously. Li et al.~\cite{li2019stereo} proposed a Stereo R-CNN-based 3D object detection model for autonomous driving by fully exploiting the sparse and dense, semantic, and geometry information in stereo imagery.

Bojarski et al.~\cite{bojarski2016end}, as part of the NVIDIA research team, proposed a new convolutional neural network (CNN) architecture for end-to-end deep learning for self-driving cars. In the new automotive application, they used convolutional neural networks (CNNs) to map the raw pixels from a front-facing camera to the steering commands for a self-driving car. Zhenye-Na ~\cite{Zhenye-Na} also implemented NVIDIA model architecture for training self-driving vehicles using Udacity's simulation environment to generate training data and test the model. In addition, Smolyakov et al. ~\cite{smolyakov2018self} explored various CNN architectures in order to obtain better results with a minimum number of parameters for predicting the steering angles to drive the car in autonomous mode in the Udacity simulation environment.

Santana and Hotz ~\cite{santana2016learning} investigated video prediction models based on autoencoders and RNNs. Instead of learning everything in an end-to-end way, they first trained the autoencoder with generative adversarial network-based cost functions to generate realistic-looking images of the road, then trained the RNN transition model in the embedded space. Behley et al. ~\cite{behleydataset} proposed three benchmark experiments based on the KITTI dataset for (i) semantic segmentation of point clouds using a single scan, (ii) semantic segmentation using multiple past scans, and (iii) semantic scene completion to understand LiDAR sequences useful for learning the environment around a self-driving car. Vora et al. ~\cite{vora2020pointpainting} proposed PointPainting: a sequential fusion method for 3D object detection inself-driving cars.  They mentioned that Point Painting works by projecting LiDAR points into the output of an image-only semantic segmentation network and appending the class scores to each point. The appended (painted) point cloud can then be fed to any LiDAR-only method. Accurate depth estimation is a key prerequisite in many robotics tasks, including autonomous driving. Guizilini~\cite{guizilini20203d} proposed a novel self-supervised monocular depth estimation method combining geometry with a new deep network, PackNet, learned only from unlabeled monocular videos. They leveraged the novel symmetrical packing and unpacking blocks to jointly learn to compress and decompress detail-preserving representations using 3D convolutions. Bertoni et al.~\cite{bertoni2019monoloco} proposed an approach to fundamentally tackle the ill-posed problem of 3D human localization from monocular RGB images for autonomous driving. They addressed the ambiguity in the task by predicting confidence intervals through a loss function based on the Laplace distribution. Gu et al. ~\cite{gu2020lstm} proposed an approach to train long short-term memory(LSTM)-based model for imitating the behavior of Waymo’s self-driving model. The proposed model was evaluated based on Mean Absolute Error (MAE) and the experimental results showed that the LSTM model outperformed several baseline models in driving action prediction. 

Image Segmentation has been an active field of research as it has a wide range of applications,  from automated disease detection to self-driving cars. Jadon~\cite{jadon2020survey} summarized existing well-known loss functions widely used for Image Segmentation and listed out the cases where their usage can help in the fast and better convergence of a model. The paper also introduced a new log-cosh dice loss function. Liao et al.~\cite{liao2022kitti} developed KITTI-360 for autonomous driving, the successor of the popular KITTI dataset. KITTI-360 is a suburban driving dataset that comprises richer input modalities, comprehensive semantic instance annotations, and accurate localization to facilitate research at the intersection of vision, graphics, and robotics. They also created a tool to label 3D scenes with bounding primitives and developed a model that transferred this information into the 2D image domain, resulting in over 150k images. Yang et al.~\cite{yang2018end} proposed a multi-task learning framework to predict the steering angle and control speed simultaneously in an end-to-end manner by taking previous feedback speeds and visual recordings as inputs.  Moghadam and Elkaim~\cite{moghadam2019hierarchical} proposed a multi-modal architecture that includes the environmental modeling of ego surrounding, trained a deep reinforcement learning (DRL) agent that yields consistent performance in stochastic highway driving scenarios, and obtained the high-level sequential commands (i.e. lane change) to send them to lower-level controllers. Deruyttere et al. ~\cite{deruyttere2019talk2car} considered a problem in an autonomous driving setting where a passenger requests an action that can be associated with an object found in a street scene. They presented the Talk2Car dataset, which was the first object referral dataset that contains commands written in natural language for self-driving cars.

\section{Methods}\label{sec:methods}
Firstly, the images and corresponding steering angles are collected from the simulator in the training mode and passed onto the CNN model as inputs. 
Secondly, the CNN model is trained for a sufficient number of epochs till the convergence in loss is achieved. Thirdly, the outputs from the CNN  model, i.e., steering angles, are passed onto the simulator in an autonomous mode. Then the car drives on its own on the selected track in the simulator.

As the model is expected to predict the steering angles, we consider the regression loss calculated using the equation below.
\begin{equation}
   \text{Regression Loss} =\frac{1}{n} \sum_{i=1}^{n}
    (y_i-\hat{y_i})^2,
\end{equation}
where $y_{i}$ is the actual steering angle and $\hat{y_i}$ is the predicted steering angle.

\subsection{Model Architecture}\label{sec:model}
\begin{figure}
\begin{center}
\includegraphics[width=0.9\textwidth]{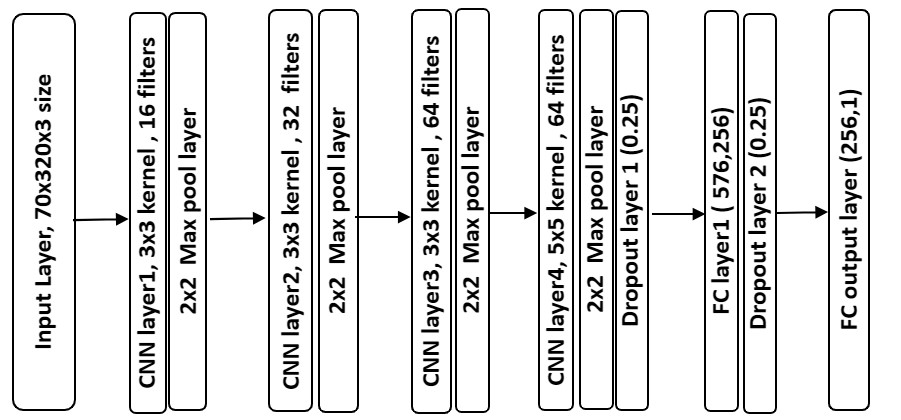}
\caption{LaksNet model architecture} \label{fig:LaksNet}
\vspace{-0.6cm}
\end{center}
\end{figure}

Our LaksNet model consists of four convolutional layers, four max-pooling layers, two dropout layers, and two fully connected layers, with one being an output layer, as shown in Fig.~\ref{fig:LaksNet}. We use 3$\times$3 kernels for the first three convolutional layers and 5$\times$5 kernels for the last convolutional layer. The first fully connected layer takes 576 input parameters and gives 256 output parameters. The second fully connected layer gives a single output because it is the final output layer. This model has a total of 274,017 training parameters.

On the other hand, the model developed using the NVIDIA architecture~\cite{bojarski2016end} consists of five convolutional layers, one dropout layer, and four fully connected layers, with one being an output layer. The total training parameters of this model are 559,419. In addition to these two model architectures, we utilized the following nine ImageNet pre-trained models to check the performance of the self-driving car in the simulation environment
(AlexNet~\cite{krizhevsky2012imagenet}, GoogleNet~\cite{szegedy2015going}, MobileNetv2~\cite{sandler2018mobilenetv2}, ResNet50~\cite{he2016deep}, SqueezeNet~\cite{iandola2016squeezenet}, DenseNet201~\cite{huang2017densely}, NasnetLarge~\cite{zoph2018learning}, ResNet101~\cite{he2016deep}, and Xception~\cite{chollet2017xception}).

\subsection{Implementation details}
Our model is implemented based on the following configurations:
\begin{enumerate}
    \item Processor: 12th Gen Intel(R) Core(TM) i7-12700H   2.30 GHz
    \item Installed RAM: 16.0 GB 
    \item System type: 64-bit operating system, x64-based processor 
    \item GPU: NVIDIA GeForce RTX 3060, 6.0 GB
\end{enumerate}

Our model is trained based on the following hyperparameters:
\begin{enumerate}
    \item Number of epochs: 50
    \item Batch size: 32
    \item Optimizer: Adam 
    \item Learning rate: 0.1
\end{enumerate}

\section{Datasets and Results}\label{sec:results and datasets}

\subsection{Datasets}\label{sec:data}
\begin{figure}
\includegraphics[width=\textwidth]{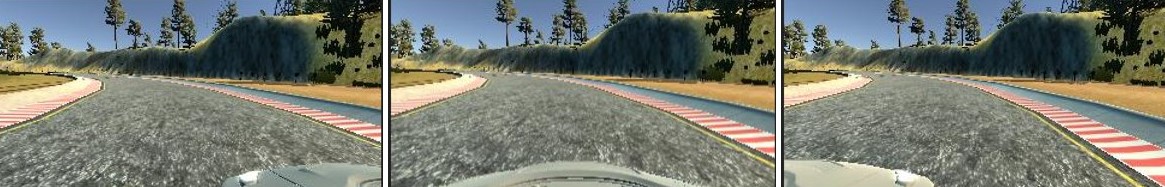}
\caption{Images from the Left, Front, and Right cameras} \label{fig:Training_images}
\end{figure}

The Udacity simulation environment has been used to generate training data. It has two different tracks that can be used for training and testing. After selecting either track1 or track2 and clicking on the record button, a folder of images is created taken using the center, right and left cameras at each instant of time. These images contain details like the portion of a car on the track, a track with borders, and the environment outside the track. A sample of images generated in training mode at one particular instant of time is shown in Fig.~\ref{fig:Training_images}.

In addition to these images, the Udacity simulation generates a drivinglog.csv file that contains seven columns. The first three columns contain the path of the images from the front, right and left cameras. Column 4 shows the steering angle values - zero corresponds to the straight direction, a positive value indicates a right turn and a negative value indicates a left turn. Columns 5, 6, and 7 indicate acceleration, deceleration, and the speed of the vehicle, respectively. Some rows of the CSV file are shown in Table.~\ref{tab:csv}.

\begin{table}
\begin{center}
\caption{Drivinglog.csv file generated by the simulator}\label{tab:csv}
\setlength{\tabcolsep}{+0.05cm}{
\begin{tabular}{lllcccc}
\hline
Column 1 & Column 2 & Column 3 & Column 4 & Column 5 & Column 6 & Column 7 \\
\hline
 center$\_$216.jpg & left$\_$216.jpg & right$\_$216.jpg & 0 & 0.3325 & 0 & 0.2858 \\
center$\_$316.jpg & left$\_$316.jpg & right$\_$316.jpg & 0 & 0.6327 & 0 & 0.8770 \\
center$\_$424.jpg & left$\_$424.jpg & right$\_$424.jpg & -0.1215 & 0.9266 & 0 & 1.8474 \\
center$\_$535.jpg & left$\_$535.jpg & right$\_$535.jpg & -0.4860 & 1 & 0 & 3.2320 \\
center$\_$637.jpg & left$\_$637.jpg & right$\_$637.jpg & -0.1827 & 1 & 0 & 4.4072 \\
center$\_$739.jpg & left$\_$739.jpg & right$\_$739.jpg & 0 & 1 & 0 & 5.5639 \\
 \hline
\end{tabular}}
\end{center}
\end{table}

We generated 33,096 images from the simulator and combined them with the training data of 97,330 images available on the GitHub page of Zhenye-Na~\cite{Zhenye-Na}. In addition, we created a separate set of images for validation purposes. 70$\times$320$\times$3 is the size of the images generated in the simulator.

\subsection{Results}\label{sec:results}
\begin{figure}
\begin{center}
\includegraphics[width=0.7\textwidth]{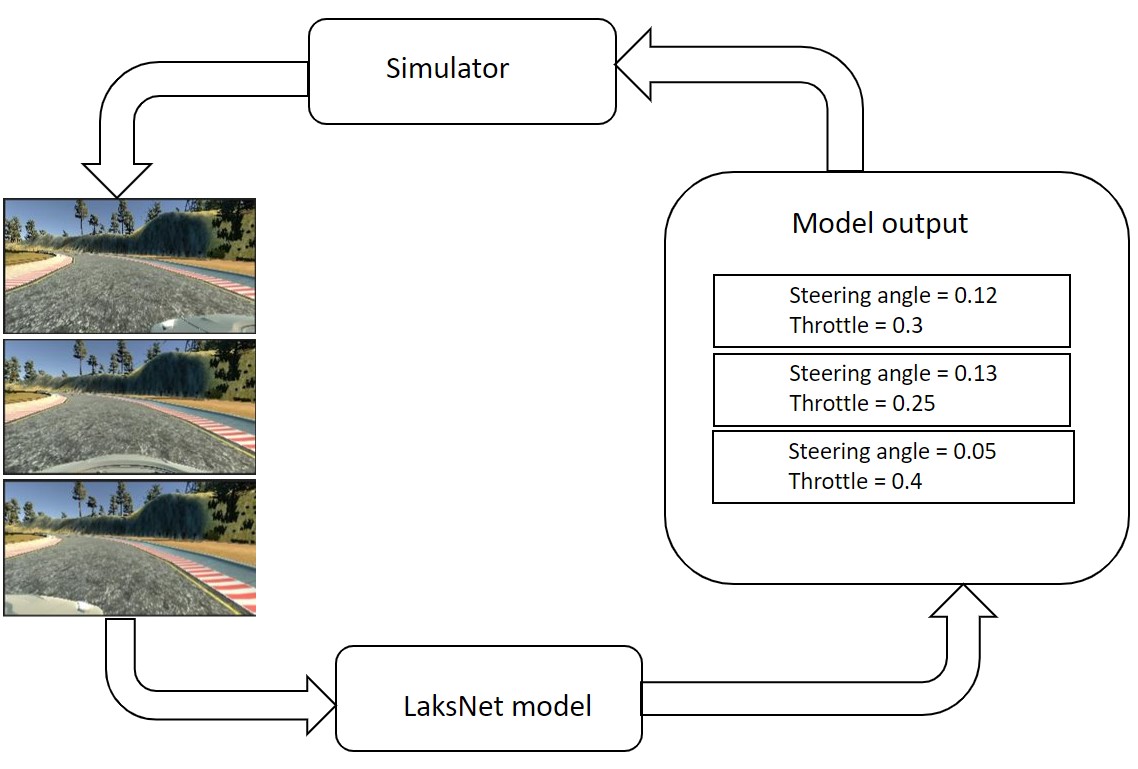}
\caption{Block diagram of testing approach in the simulator} \label{fig:Testing_approach}
\vspace{-0.6cm}
\end{center}
\end{figure}

The block diagram shown in Fig.~\ref{fig:Testing_approach} highlights the testing approach of the model to drive the car in autonomous mode in the simulation environment. The training images and the respective steering angles generated from the Udacity simulator are fed into the CNN model for training. After training for a required number of epochs, the model could predict steering angle values that are passed into the simulator. We notice that the car is able to drive on its own on the track in the simulation environment in autonomous mode.

\begin{figure}
\begin{center}
\includegraphics[width=0.9\textwidth]{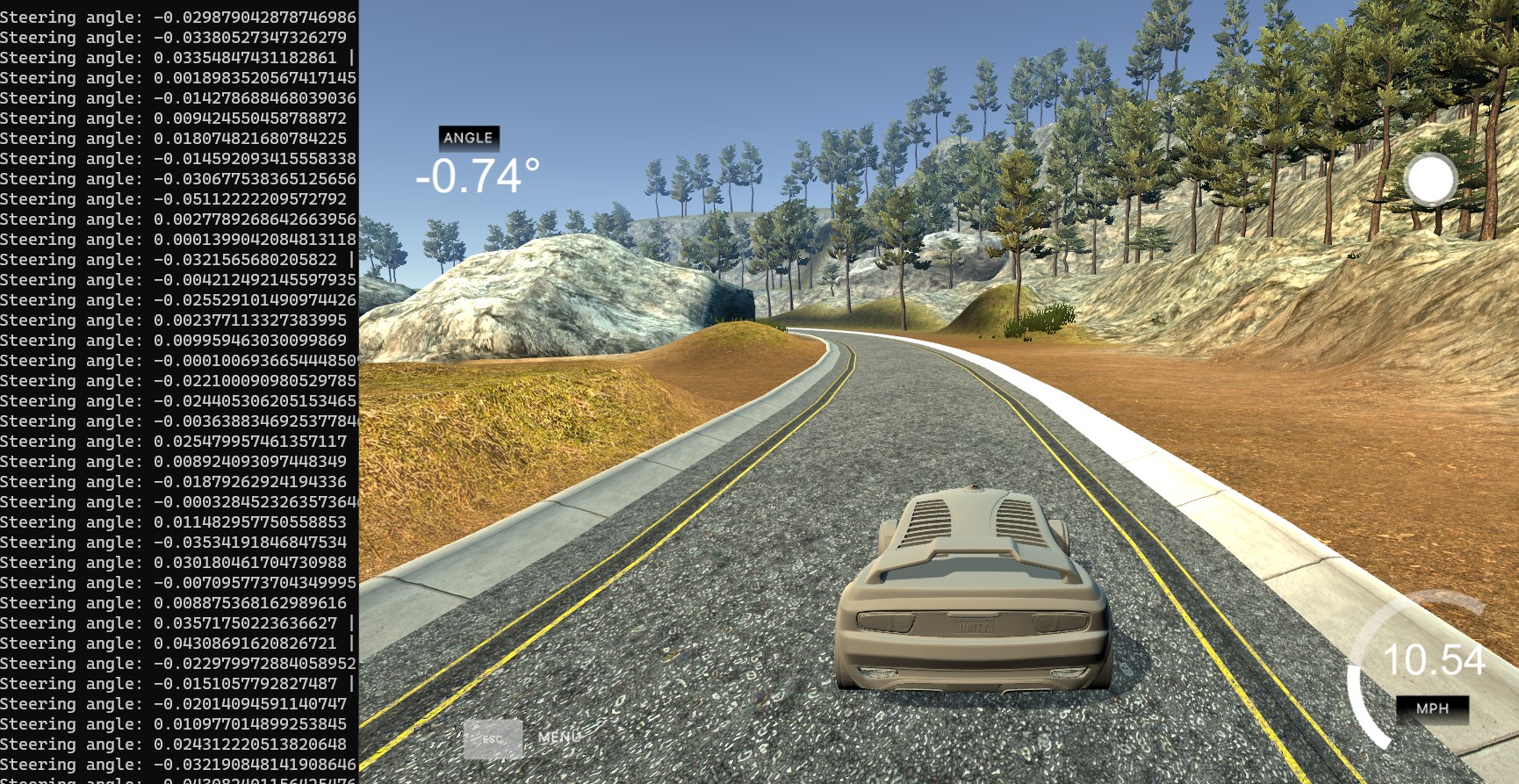}
\caption{A snapshot of steering angles predicted by the model in the terminal  and simulator window in autonomous mode} \label{fig:Autonomous_mode}
\end{center}
\end{figure}

During the testing stage, we notice that the steering angles on the terminal as well as the car moved in the simulator window at the same time. A snapshot of this is shown in Fig.~\ref{fig:Autonomous_mode}. We  can also save the images taken by the front camera during autonomous mode.

\begin{table}
\begin{center}
\caption{Car run time(in secs) on the track in the simulator for each model}\label{tab:run time}
\setlength{\tabcolsep}{+0.3cm}{
\begin{tabular}{lcc}
\hline
Model & Run time(in secs)\\
\hline
AlexNet~\cite{krizhevsky2012imagenet} & 50 \\
GoogleNet~\cite{szegedy2015going}  & 50 \\
MobileNet~\cite{sandler2018mobilenetv2} & 8 \\ 
ResNext50~\cite{he2016deep}  & 3 \\
SqueezeNet~\cite{iandola2016squeezenet}  & 12 \\
DenseNet201~\cite{huang2017densely} & 7 \\ 
NasnetLarge~\cite{zoph2018learning} & 4 \\
ResNet101~\cite{he2016deep} & 14 \\
Xception~\cite{chollet2017xception} & 5 \\
NVIDIA~\cite{bojarski2016end} & 120 \\ 
\hline 
\textbf{LaksNet} & \textbf{150} \\
\hline    
\end{tabular}}
\end{center}
\end{table}

After training each of the models, we obtain the results shown in Table.~\ref{tab:run time}. The run time (in seconds) refers to the duration of which the car runs at an average speed of 10 miles per hour without going off the track. These results highlight that the LaksNet model has a better performance than that of NVIDIA and other pre-trained models. Table.~\ref{tab:angle} shows the actual steering angles, predicted steering angles, and MSE (Mean Square Error) of all models for 30 test images generated separately from the Udacity simulator. Though the MSE of the LaksNet model is larger than that of the NVIDIA model, it could drive the car on track for more time. This might be due to the fact that it did not overfit the data.

\begin{table}
\tiny
\begin{center}
\caption{Predicted steering angles and MSE by each model for 30 test images (ANet: Alexnet; GNet: Googlenet; MNet: Mobilenet; R50: Resnext50; SqNet: Squeezenet; D201: Densenet201; NasL: Nasnetalarge; R101: Resnet101) }\label{tab:angle}
\begin{tabular}{|llllllllllll|}
\hline
Actual angles & \textbf{LaksNet} & NVIDIA & ANet & GNet & MNet & R50 & SqNet & D201 & NasL & R101 & Xception \\
\hline
 -0.341 & -0.348 & -0.079 & -0.133 & -0.288 & -0.122 & -0.040 & 0.000 & 0.053 & 3.315 & -0.039 & -0.005 \\
0.000 & -0.355 & -0.082 & -0.033 & -0.123 & 0.078 & -0.200 & 0.000 & 0.025 & 3.294 & 0.110 & 0.026			 \\ 
0.000 & -0.363 & -0.098 & -0.289 & -0.201 & -0.100 & -0.216 & 0.000 & 0.014 & 3.458 & 0.018 & 0.009		 \\
 -0.072 & -0.375 & -0.186 & -0.053 & -0.144 & -0.202 & -0.121 & 0.000 & 0.002 & 3.507 & 0.001 & 0.084 \\ 
-0.418 & -0.355 & -0.079 & -0.035 & -0.148 & -0.073 & -0.109 & 0.000 & 0.037 & 3.459 & -0.073 & -0.029		 \\ 
 -0.116 & -0.350 & -0.094 & -0.206 & -0.105 & -0.089 & 0.015 & 0.000 & -0.006 & 3.563	& -0.009 & 0.001	 \\
 0.000 & -0.325 & -0.032 & -0.225 & -0.191 & -0.108 & -0.126 & 0.014 & 0.032 & 3.421 & -0.190 & -0.005			 \\
 0.000 & -0.251 & -0.060 & -0.169 & -0.121 & -0.064 & -0.200 & 0.000 & -0.032 & 3.511 & -0.152 & 	-0.018		 \\ 
 0.000 & -0.280 & -0.030 & 0.001 & -0.311 & -0.290 & -0.070 & 0.0229 & 0.032 & 3.576 & -0.088 & -0.037			 \\
 0.000 & -0.375 & -0.036 & -0.197 & -0.014 & -0.109 & -0.112 & 0.025 & 0.014	& 3.652 & -0.074 & -0.034		 \\
0.000 & -0.360 & -0.032 & -0.297 & -0.210 & -0.275 & -0.138 & 0.000 & -0.017 & 3.742 & -0.108 & -0.008			\\
0.000 & -0.383 & -0.068 & -0.023 & -0.156 & -0.091 & -0.112 & 0.000 & 0.041	& 3.632 & -0.030 & 0.681	\\
0.000 & -0.417 & -0.024 & -0.032 & -0.094 & -0.093 & -0.318 & 0.000 & -0.006 & 3.742 & -0.182 & 0.004	\\
-0.071 & -0.464 & -0.059 & -0.130 & -0.011 & -0.101 & -0.017 & 0.000 & -0.075 & 3.745	& 0.001	& 0.029	\\
-0.391 & -0.395 & -0.064 & 0.001 & -0.073 & -0.312 & -0.124 & 0.000 & -0.041 & 3.978 & -0.103 & -0.013		\\
-0.279 & -0.402 & -0.055 & -0.057 & -0.123 & -0.047 & -0.072 & 0.000 & 0.029 & 3.722 & -0.065	& -0.049	\\
0.000 & -0.264 & -0.045 & -0.044 & -0.038 & -0.057 & 0.006 & 0.000 & 0.144 & 3.901 & -0.105 & -0.012	\\
0.000 & -0.219 & -0.029 & -0.020 & -0.006 & -0.071 & -0.056 & 0.000 & 0.000 & 3.840 & -0.151 & 0.085			\\
0.000 & -0.206 & -0.049 & -0.147 & -0.105 & -0.318 & -0.071 & 0.000 & 0.066 & 3.818 & 0.015 & 0.003			\\
0.000 & -0.238 & -0.048 & -0.065 & -0.123 & -0.028 & -0.023 & 0.000 & 0.076 & 3.703 & -0.020 & -0.001			\\
0.000 & -0.237 & -0.056 & -0.012 & -0.119 & -0.026 & -0.040 & 0.000 & 0.053	& 3.794 & 0.101 & -0.003	\\
0.000 & -0.280 & -0.042 & -0.228 & -0.023 & -0.039 & -0.045 & 0.000 & 0.092 & 3.583 & 0.015 & 0.001			\\
0.000 & -0.278 & -0.044 & 0.034 & -0.069 & -0.054 & -0.050 & 0.000 & 0.011 & 3.625 & 0.022 & -0.014		\\
0.000 & -0.251 & -0.036 & -0.135 & -0.116 & -0.038 & -0.239 & 0.000 & 0.051	& 3.692 & -0.057 & 0.011		\\
0.000 & -0.335 & -0.053 & -0.007 & -0.075 & -0.049 & -0.231 & 0.000 & 0.083	& 3.771 & 0.015 & -0.026	\\
0.000 & -0.305 & -0.040 & -0.184 & -0.178 & -0.081 & -0.074 & 0.000 & -0.026 & 3.762 & -0.037 & -0.013		\\
0.000 & -0.483 & -0.051 & -0.068 & -0.151 & -0.181 & -0.096 & 0.000 & 0.088 & 3.651 & -0.052 & -0.014		\\
0.000 & -0.342 & -0.051 & -0.077 & -0.167 & -0.089 & -0.092 & 0.000 & 0.116 & 3.557 & 0.026 & -0.019			\\
0.000 & -0.381 & -0.074 & -0.244 & -0.214 & -0.131 & -0.010 & 0.000 & -0.027 & 3.518 & -0.042 & -0.006		\\
0.000 & -0.345 & -0.083 & -0.151 & -0.108 & -0.063 & -0.038 & 0.000 & -0.113 & 3.589 & -0.093 & -0.018		\\
\hline
\textbf{MSE} & 0.091 & 0.014 & 0.031 & 0.023 & 0.021 & 0.047 & 0.066 & 0.023 & 13.682 & 0.020 & 0.033 \\
\hline

\end{tabular}
\end{center}
\end{table}

\section{Discussion}\label{sec:disc}
As we are inspired by the NVIDIA architecture for end-to-end learning of a self-driving car, we first train this model with our dataset and notice that the car ran properly on the track in the simulator for 120 seconds in autonomous mode and deviated from the track. Then, we decide to check the performance of the available pre-trained models for this purpose. Surprisingly, these complex models have shown poor performance when compared to the NVIDIA model. This might be due to the fact that all these pre-trained models are designed for classification tasks to identify a particular category instead of predicting numerical values. Our project is a regression problem as we predict continuous numerical values (steering angles). Besides, similar images can have different angles, which also confuses these complex models. Among these models, Alexnet and Googlenet show better results compared to other pre-trained models because these two are simpler models in terms of the number of convolutional layers.

As the pre-trained models did not meet the expectations set by the NVIDIA model, we tried to build our own CNN models that are more competent. One of our CNN models, with only four convolutional layers and two fully connected layers, could drive the car on the track properly for 150 seconds, more than that of the NVIDIA model. To arrive at this model, we have built many models and checked their performances.

\subsection{Hyper-parameter tuning}\label{sec:hyper}
We first explored a CNN model with seven convolutional layers using 3$\times$3 filters. We initially thought that adding more convolutional layers would make the model achieve better results, but this model only drove properly for 20 seconds. We observed that adding more layers would extract features up to a certain extent and eventually overfit the data. So, we built a few more models with five and six convolutional layers using bigger kernel sizes of 5$\times$5 and 7$\times$7. We observed that the model with five convolutional layers using 5x5 filters and the model using a combination of 7$\times$7 and 5$\times$5 filters showed a major improvement in their performance by driving the car properly on the track for 90 seconds. We then developed another model with 3$\times$3 convolutional layers using 7$\times$7 filters, but it could drive the car properly for just 50 seconds. When we reduced the filter size to 3$\times$3 in this model, it could drive the car properly for 120 seconds, meeting the expectations of the NVIDIA model. Therefore, we discovered that models with a smaller number of layers and using smaller filter sizes could generate better results, as shown below in Table.~\ref{tab:ab}.

\begin{table}
\begin{center}
\caption{Run time of the car for various model configurations}\label{tab:ab}
\setlength{\tabcolsep}{+0.3cm}{
\begin{tabular}{ll}
\hline
Model configuration &  Run time(in secs)\\
\hline
 Seven layers, all 3$\times$3 filters & 20 \\
 Five layers, all 5$\times$5 filters &  90 \\ 
 Five layers, three 7$\times$7 filters and two 5$\times$5 filters & 90 \\
 Three layers, all 7$\times$7 filters & 50 \\
 Three layers, all 3$\times$3 filters & 120 \\ 
 \hline
\end{tabular}}    
\end{center}
\end{table}

In an attempt to achieve better results than that of the NVIDIA model, we conducted experiments with various models using 3 or 4 convolutional layers with 3$\times$3 or 5$\times$5 or a combination of 3$\times$3 and 5$\times$5 filters. Finally, our LaksNet model with four convolutional layers, using 3$\times$3 filters for the first three layers and a 5$\times$5 filter for the last layer, achieves the best results in driving the car properly for 150 seconds. In this model, we only use two fully connected layers, with one being an output layer. When we test this model with 3 or 4 fully connected layers, the model overfitted and did not yield better results.

In the LaksNet model, we add a max pooling layer after each convolutional layer  to reduce the height and width of the images. Another alternative way is to use an average pooling layer. In the fully connected block, we add a dropout layer after the last convolutional layer and the first fully connected layer  to avoid overfitting. We apply the ReLU activation function for all convolutional and fully connected layers, except the output layer, to fire neurons. The other suitable alternative to ReLU is the ELU activation function.

Regarding the loss function, we apply the MSE loss function because the output is one value, and the loss is measured by calculating the difference between the actual and predicted steering angle values. Accordingly, weights will be adjusted during backpropagation and this process continues based on the number of epochs. In addition to the data augmentation process, we normalize the data to avoid saturation so that gradients would work better for the model training. We also perform random rotations on the images and crop the images as well. This is required for better model training. When generating the training data from the simulator, one has to ensure that the car always moves in the center of the track while driving manually, otherwise, the training data will be erroneous, thus reducing the model performance during predictions. In addition, a large training dataset is required for better learning of the model in terms of turns on the track and surrounding environment. 

\section{Conclusion}\label{sec:concl}
In this paper, we first train the NVIDIA model and other pre-trained models with 134K images generated from the Udacity simulator and demonstrate that the NVIDIA model outperforms the pre-trained models in terms of car run time on the track. Secondly, we build the LaksNet model with only four convolutional layers and two fully connected layers using a smaller number of training parameters. While the NVIDIA model is able to drive the car for 120 seconds before going off-track, LaksNet shows significant improvements by driving the car for 150 seconds. Our future work will focus on building a model that could run the car for a longer time without going off the track under the Udacity simulator. In addition, we could predict acceleration values along with steering angles and ensure that the car will always run in the middle of the track.




%
%
%
\bibliographystyle{splncs04}
\bibliography{mybibliography}
%




\end{document}